\pdfoutput=1

\documentclass[11pt]{article}

\usepackage{acl}

\usepackage{times}
\usepackage{latexsym}

\usepackage[T1]{fontenc}

\usepackage[utf8]{inputenc}

\usepackage{microtype}

\usepackage{inconsolata}

\usepackage{graphicx}

\title{dzStance at StanceEval2024: Arabic Stance Detection based on Sentence Transformers}

\author{Mohamed Lichouri \\
  LCPTS, FGE-USTHB \\
  Algiers-ALGERIA \\
  \texttt{mlichouri@usthb.dz} \\\And
  Khaled Lounnas \\
  CRSTDLA \\
  Algiers-ALGERIA \\
  \texttt{k.lounnas@crstdla.dz} \\\AND
  Khelil Rafik Ouaras, Mohamed Abi and Anis Guechtouli \\
  Algiers 01 University\\
  Algiers-ALGERIA \\
    \texttt{\{kh.ouaras, m.abi, a.guechtouli\}@univ-alger.dz}
 \\}

\begin{document}
\maketitle
\begin{abstract}
This study compares Term Frequency-Inverse Document Frequency (TF-IDF) features with Sentence Transformers for detecting writers' stances—favorable, opposing, or neutral—towards three significant topics: COVID-19 vaccine, digital transformation, and women empowerment. Through empirical evaluation, we demonstrate that Sentence Transformers outperform TF-IDF features across various experimental setups. Our team, dzStance, participated in a stance detection competition, achieving the 13th position (74.91\%) among 15 teams in Women Empowerment, 10th (73.43\%) in COVID Vaccine, and 12th (66.97\%) in Digital Transformation. Overall, our team's performance ranked 13th (71.77\%) among all participants. Notably, our approach achieved promising F1-scores, highlighting its effectiveness in identifying writers' stances on diverse topics. These results underscore the potential of Sentence Transformers to enhance stance detection models for addressing critical societal issues.
\end{abstract}

\section{Introduction}
\label{intro}
Stance detection is a pivotal task in Natural Language Processing (NLP) that involves determining the position or attitude expressed in a text regarding a specific topic or entity. This task is crucial for a variety of applications, including sentiment analysis, opinion mining, and social media monitoring, where understanding public sentiment and opinion is essential \cite{alturayeif2022mawqif}. With the exponential growth of user-generated content on social media and online news platforms, there is a pressing need to develop sophisticated tools that can analyze and interpret the myriad of perspectives present in these texts.

The Mawqif 2022 shared task, an initiative focused on Arabic stance detection, addresses this need by challenging participants to detect stances towards three contemporary topics: COVID-19 vaccine, digital transformation, and women empowerment. This task is significant not only because it addresses critical societal issues but also because it highlights the complexities of processing Arabic text, which is characterized by its rich morphology, diverse dialects, and intricate syntax \cite{alturayeif2022mawqif}.

Traditional methods for stance detection have largely relied on feature extraction techniques such as Term Frequency-Inverse Document Frequency (TF-IDF). These methods transform textual data into a numerical format, allowing machine learning models to process and analyze the text. Our initial experiments leverage TF-IDF features due to their simplicity and proven effectiveness in various text classification tasks \cite{abbas2019st, lichouri2020speechtrans}.

Despite their utility, TF-IDF-based methods have limitations, particularly in capturing the deeper semantic relationships and contextual nuances within text. To address these limitations, recent research has increasingly focused on deep learning techniques. Long Short-Term Memory (LSTM) networks have demonstrated their ability to handle sequential data and capture dependencies in text, which are crucial for understanding stances \cite{alturayeif2023systematic, lai2020multilingual}.

The advent of transformer-based models, such as BERT and Sentence Transformers, has revolutionized the field of NLP by offering robust methods to capture semantic and contextual information in text. These models utilize self-attention mechanisms to understand the relationships within text, making them particularly effective for nuanced tasks like stance detection \cite{reimers-2019-sentence-bert, aldayel2021stance}. Transformers have set new benchmarks in various NLP tasks by leveraging their ability to generate dense, context-aware representations of text, facilitating a deeper understanding of the underlying meaning and intent.

Several studies have explored the use of BERT and other transformers for stance detection, demonstrating their superiority over traditional methods in capturing subtle and complex stances in text. For example, Alshahrani et al. \cite{aldayel2021stance} showed that BERT-based models outperform traditional approaches in detecting stances in English social media texts. However, these models often require extensive computational resources and fine-tuning, posing challenges for their application in resource-limited settings \cite{alturayeif2023systematic}.

Building upon these advancements, our study explores a comparative analysis between traditional feature extraction methods and modern deep learning approaches for Arabic stance detection. In our experiments, we employ both TF-IDF and Sentence Transformers to detect stances towards the selected topics. Our participation in the Mawqif 2022 shared task allowed us to evaluate these methodologies rigorously, where our team, dzStance, achieved competitive results across the three topics, demonstrating the effectiveness of our approaches.

In the following sections, we will delve deeper into our methodology and findings. Section \ref{data} provides an overview of the Mawqif dataset used in our study. Section \ref{system} describes our approach to stance detection, including the feature extraction techniques and model architectures we employed. Section \ref{res} presents our experimental results and discusses their implications. Finally, Section \ref{conc} concludes the paper by summarizing the key insights and contributions of our work.

\section{Dataset Description}
\label{data}

The \textbf{Mawqif dataset} \cite{alturayeif2022mawqif}, utilized in the \textbf{StanceEval 2024 shared task} \cite{StanceEval2024}, serves as a crucial resource for advancing natural language processing (NLP) in the domain of stance detection. This dataset comprises over 4,000 annotated text samples that encapsulate diverse stances—\textit{favorable}, \textit{opposing}, or \textit{neutral}—on pertinent topics such as COVID-19 vaccine, digital transformation, and women empowerment.

The significance of the Mawqif dataset lies in its ability to provide a comprehensive view of how different opinions and attitudes are expressed in Arabic text. This makes it invaluable for researchers who aim to evaluate and enhance stance detection models. By leveraging such a dataset, one can explore and refine models to better understand and process nuanced stances within varied contexts.

Table \ref{tab:data_split} offers a detailed breakdown of the dataset, illustrating the distribution of tweets across the specified topics. The dataset comprises a total of 3,502 tweets, with the distribution as follows:

\begin{table*}[h]
\centering
\caption{Data statistics of MAWQIF dataset.}
\label{tab:data_split}
\begin{tabular}{lcccc}
\hline
\textbf{Target} & \textbf{\#Tweets} & \textbf{\#Favor} & \textbf{\#Against} & \textbf{\#None} \\ \hline
COVID-19 Vaccine & 1167 & 508 & 507 & 152  \\
Digital Transformation & 1145 & 879 & 142  & 22 \\
Women Empowerment & 1190 & 761 & 371 & 59 \\
All & 3502 & 2154 & 1020 & 332  \\ \hline
\end{tabular}
\end{table*}

A closer look at the data statistics reveals notable class imbalances. For instance, the COVID-19 vaccine category includes nearly equal proportions of favorable (43.53\%) and opposing (43.48\%) tweets, with a smaller fraction being neutral (12.85\%). In contrast, the digital transformation topic shows a predominance of favorable stances (76.77\%), with fewer opposing (12.41\%) and neutral (1.92\%) tweets. Similarly, the women empowerment category also leans heavily towards favorable stances (63.95\%), followed by opposing (31.18\%) and neutral (4.96\%) tweets.

Such imbalances can pose significant challenges for model training and evaluation. Models trained on datasets with skewed class distributions may become biased towards the majority classes, leading to suboptimal performance on minority classes. Therefore, addressing these imbalances is critical to ensure the development of robust and fair models. Techniques like data resampling, class weighting, and the use of advanced algorithms capable of handling imbalance are essential strategies to mitigate these effects and enhance overall model performance.

By understanding and leveraging the characteristics of the Mawqif dataset, researchers can effectively tackle the complexities of stance detection, contributing to the broader field of Arabic NLP and enabling more accurate and nuanced analysis of opinions and attitudes expressed in text.

\section{Proposed System}
\label{system}
In our proposed system \footnote{https://github.com/licvol/dzStanceEval\_2024}, we explore two distinct methodologies for feature extraction: a weighted union of TF-IDF features \cite{lichouri2023usthb} and Sentence Transformers. These techniques offer complementary advantages, leveraging the strengths of traditional feature representation and cutting-edge deep learning architectures.

\begin{table*}[!h]
\centering
\begin{tabular}{|r|l|l|l|l|l|}
\hline
Id & Model               & Text Feat                                               & Configuration                                  & Other                              & F1-score \\ \hline
1                      & LSVC      & ngram\_range=(1,4),(1,4),(1,4)                             & C=4                                             & tw1              & 60.57 \\ \hline
2                      & LSVC      & ngram\_range=(1,1),(1,1),(1,1)                             & C=4                                             & tw1              & 61.15 \\ \hline
3                      & LSVC      & ngram\_range=(1,5),(1,5),(1,5)                             & C=4                                             & tw1              & 61.27 \\ \hline
4                      & LSVC      & ngram\_range=(1,7),(1,7),(1,7)                             & C=4                                             & tw1              & 61.32 \\ \hline
5                      & LSVC      & ngram\_range=(1,10),(1,10),(1,10)                          & C=4                                             & tw1              & 61.9  \\ \hline
6                      & LSVC      & ngram\_range=(1,2),(1,2),(1,2)                             & C=4                                             & tw1              & 62.1  \\ \hline
7                      & LSVC      & ngram\_range=(1,3),(1,3),(1,3)                             & C=4                                             & tw1              & 63.82 \\ \hline
8                      & LSVC      & ngram\_range=(1,8),(1,8),(1,8)                             & C=4                                             & tw1              & 63.92 \\ \hline
9  & LSVC      & ngram\_range=(1,9),(1,9),(1,9)                             & C=4                                             & tw1              & 64.66 \\ \hline
10 & LSVC      & ngram\_range=(1,6),(1,6),(1,6)                             & C=4                                             & tw1              & 66.2  \\ \hline \hline
11 & LR & XLM-RoBERTa & \begin{tabular}[c]{@{}l@{}}max\_iter=1000 \\ multi\_class='multinomial'\\ solver='lbfgs'\end{tabular}  & na, re & \textbf{68.48}                        \\ \hline
\end{tabular}
\caption{Obtained F1-score in the development phase. tw1:=\{0.85,0.85,0.65\}, na:=normalize\_arabic and re:=replace\_emojis}
\label{tab:DZStance}
\end{table*}

\begin{description}
    \item[Experiment 1] : Traditional Machine Learning with TF-IDF Features
    In our first experiment, we focused on extracting features using the Term Frequency-Inverse Document Frequency (TF-IDF) approach, which is widely used in text classification tasks. We utilized scikit-learn's \texttt{FeatureUnion} module to combine different TF-IDF features, capturing both character-level and word-level information \cite{lichouri-etal-2021-arabic}.
    
    \begin{itemize}
        \item \textbf{N-gram Range:} We experimented with various n-gram ranges to understand their impact on model performance. The ranges included:
        \begin{itemize}
            \item \texttt{(1,1)}: Unigrams
            \item \texttt{(1,2)}: Unigrams and bigrams
            \item \texttt{(1,3)}: Up to trigrams
            \item \texttt{(1,4)}: Up to 4-grams
            \item \texttt{(1,5)}: Up to 5-grams
            \item \texttt{(1,6)}: Up to 6-grams
            \item \texttt{(1,7)}: Up to 7-grams
            \item \texttt{(1,8)}: Up to 8-grams
            \item \texttt{(1,9)}: Up to 9-grams
            \item \texttt{(1,10)}: Up to 10-grams
        \end{itemize}

        \item \textbf{Weighting:} To further enhance the TF-IDF features, we incorporated weighting schemes. The weights were varied from 0.1 to 1.0 in steps of 0.1 to determine the optimal balance for capturing the intricacies of Arabic text. The best-performing weight, denoted as \texttt{tw1}, was identified and applied consistently across subsequent experiments (see Table \ref{tab:DZStance}).

        \item \textbf{Classifier:} For classification, we employed the Linear Support Vector Classifier (LSVC) with a regularization parameter set to \(C=4\). This choice was based on its ability to handle high-dimensional feature spaces effectively, which is crucial when dealing with the extensive n-gram features produced by TF-IDF.
    \end{itemize}

    This combination of weighted TF-IDF features and LSVC forms the baseline for our stance detection system, aiming to capture both the surface-level and deeper linguistic patterns in Arabic text.

    \item[Experiment 2]: Leveraging Pre-trained Language Models (PLMs)
    In the second experiment, we explored the use of advanced pre-trained language models (PLMs) to enhance stance detection capabilities further. These models are pre-trained on vast amounts of text data and are adept at generating rich semantic representations of words and sentences.
    
    \begin{itemize}
        \item \textbf{Sentence Embeddings:} We utilized Sentence Transformers, specifically the \textbf{xlm-r-bert-base-nli-stsb-mean-tokens} model, which excels at producing dense vector embeddings that encapsulate the overall meaning of sentences. These embeddings were used as input features for the classification task.

        \item \textbf{Classifier:} For classification, we opted for a Logistic Regression (LR) model configured with the following hyperparameters:
        \begin{itemize}
            \item \texttt{max\_iter=1000}: This parameter sets the maximum number of iterations for the solver to converge.
            \item \texttt{multi\_class='multinomial'}: This setting enables the classifier to handle multiple classes simultaneously, which is crucial for stance detection where multiple stance labels exist.
            \item \texttt{solver='lbfgs'}: The solver used for optimization, chosen for its efficiency in handling multiclass logistic regression problems.
        \end{itemize}

        This approach leverages the powerful representations learned by Sentence Transformers, which are adept at capturing semantic nuances and contextual relationships within the text. By integrating these embeddings with a logistic regression classifier, we aim to improve the model's ability to discern subtle stance indicators in Arabic text.
    \end{itemize}
\end{description}

Combining these methodologies, our system aims to balance the strengths of traditional feature extraction with the advanced capabilities of modern pre-trained models. This hybrid approach is designed to address the linguistic complexity and variability of Arabic, providing robust stance detection across different contexts.

\section{Results and Discussion}
\label{res}

This section evaluates the performance of our stance detection system on the test set using the F1-score as the primary metric. We conducted two sets of experiments to compare the effectiveness of different feature extraction techniques and model configurations. Following these, we analyze our competitive performance in the stance detection challenge.

\subsection{Baseline Experiment}

To establish a benchmark, we employed a simple approach using Term Frequency-Inverse Document Frequency (TF-IDF) representation with unigram (1-gram) features and a Linear Support Vector Classifier (LSVC) with a linear kernel. This baseline model achieved an F1-score of 64.34\%, providing a reference point for comparing the performance of more advanced models and feature combinations.

\subsection{Experiment 1: Weighted Union of TF-IDF Features}

In this experiment, we explored the impact of various n-gram ranges on the performance of the LSVC model. We employed the FeatureUnion module to create a weighted combination of TF-IDF features with different n-gram lengths. The weights were varied systematically from 0.1 to 1.0 to optimize feature importance. The optimal weight configuration (0.85, 0.85, 0.65) from this tuning was then used across all n-gram experiments.

We tested a range of n-grams from single-word (1-gram) up to ten-word sequences, examining the effects of character-level and word-boundary-aware features. The results show a consistent trend: the F1-score improves as the n-gram range increases, reaching a peak of 66.20\% with six-grams (ngram\_range=(1,6)). This suggests that incorporating up to six-word sequences captures essential context and relationships, enhancing the model's performance in stance detection tasks.

Interestingly, the performance slightly declines for n-grams longer than six-grams (e.g., ngram\_range=(1,7) or ngram\_range=(1,8)), possibly due to the introduction of noise or redundant information. These findings indicate that while expanding n-gram ranges can enrich the feature set, overly long sequences may adversely affect model accuracy.

\subsection{Experiment 2: Sentence Transformers}

The second experiment employed pre-trained language models to generate rich sentence embeddings. We used Sentence Transformers, specifically the '\textbf{xlm-r-bert-base-nli-stsb-mean-tokens}' model, to create embeddings that encapsulate the semantic meaning of each sentence. These embeddings were then fed into a Logistic Regression (LR) classifier for stance detection.

Configured with default hyperparameters—max\_iter=1000, multi\_class='multinomial', and solver='lbfgs'—and additional text preprocessing steps such as normalization and emoji replacement, the LR model achieved an F1-score of 68.48\%. This score surpasses the highest result obtained in the TF-IDF-based LSVC experiments, highlighting the effectiveness of Sentence Transformers in capturing semantic relationships within text.

\subsection{Competitive Performance Analysis}

Our team, dzStance, participated in the StanceEval shared task, where the competition focused on detecting stances in various topical domains. Our performance across different topics was as follows:
- **Women Empowerment**: We achieved the 13th position with an F1-score of 74.91\% among 15 participating teams.
- **COVID-19 Vaccine**: We ranked 10th, securing a 73.43\% F1-score.
- **Digital Transformation**: We placed 12th with a 66.97\% F1-score.
- **Overall Performance**: Combining all categories, dzStance ranked 13th overall with an F1-score of 71.77\%.

These results reflect our system's ability to handle complex stance detection tasks, particularly in the context of the nuanced and diverse opinions expressed in the dataset. Despite the competitive nature of the task, our approach demonstrated robustness across different domains, indicating its potential for broader applications in stance detection.

\subsection{Analysis and Insights}

Our experiments underscore the importance of feature engineering and model selection in stance detection tasks. For the LSVC model, carefully selecting and combining n-grams up to six words proved most effective. This approach aligns with the need to capture both local and contextual information in Arabic text, characterized by its rich morphology and varying dialects.

On the other hand, the use of Sentence Transformers and Logistic Regression provided a significant performance boost. This suggests that leveraging pre-trained embeddings, which encode comprehensive semantic information, can substantially enhance the ability of stance detection models to interpret complex texts.

Overall, our competitive performance highlights areas for improvement but also demonstrates the potential of our methodologies. Future work could explore further fine-tuning of pre-trained models or combining TF-IDF and embedding-based approaches to harness the strengths of both methods, potentially leading to even greater improvements in stance detection performance.

\section{Conclusion}
\label{conc}
In this paper, we introduced dzStance, our solution to the StanceEval 2024 shared task on Arabic stance detection. Leveraging Sentence Transformers in conjunction with Logistic Regression, our approach achieved competitive results with an overall average F1-score of 71.77\%. This performance positioned us 13th among all participating teams, highlighting the effectiveness of advanced embedding techniques and robust classification algorithms in handling the complexities of Arabic stance detection. The success of our approach underscores the significance of pre-trained models like Sentence Transformers for capturing nuanced semantic relationships within Arabic text across diverse topics.

Looking ahead, further investigation into why the Logistic Regression model outperformed traditional methods such as Linear Support Vector Classification (LSVC) could yield insights through deeper hyperparameter tuning and broader evaluation on varied datasets. Additionally, exploring hybrid approaches that integrate TF-IDF features with advanced embedding models may offer enhanced model robustness and accuracy in Arabic natural language processing tasks. By making our code and methodologies openly accessible, we aim to foster reproducibility and encourage ongoing advancements in Arabic stance detection research, paving the way for more sophisticated and effective models in the future.

\bibliography{custom}

\end{document}